\documentclass[letterpaper]{article} 
\usepackage{aaai23}  
\usepackage{times}  
\usepackage{helvet}  
\usepackage{courier}  
\usepackage[hyphens]{url}  
\usepackage{graphicx} 
\usepackage{natbib}  
\usepackage{caption} 
\usepackage[table]{xcolor}
\usepackage{adjustbox}
\usepackage{subcaption}
\usepackage{amsmath}
\usepackage{amsthm}
\usepackage{booktabs}
\usepackage{algorithm}
\usepackage{multirow}
\usepackage{algorithmic}
\usepackage{amsmath,amssymb}
\usepackage{tabularx,ragged2e}
\newcolumntype{C}{>{\Centering\arraybackslash}X}

\DeclareMathOperator{\EX}{\mathbb{E}}

\usepackage{algorithm}
\usepackage{algorithmic}

\usepackage{newfloat}
\usepackage{listings}
\DeclareCaptionStyle{ruled}{labelfont=normalfont,labelsep=colon,strut=off} 
\floatstyle{ruled}
\newfloat{listing}{tb}{lst}{}
\floatname{listing}{Listing}
\title{I Know Therefore I Score: Label-Free Crafting of Scoring Functions using Constraints Based on Domain Expertise}
\author {
    Ragja Palakkadavath\textsuperscript{\rm 1,\thanks{This work was done while at TCS Research, Pune}} 
    Sarath Sivaprasad\textsuperscript{\rm 2, \footnotemark[1]} 
    Shirish Karande\textsuperscript{\rm 3} 
    Niranjan Pedanekar\textsuperscript{\rm 3}}
    
\affiliations{
    \textsuperscript{\rm 1}Deakin University 
    \textsuperscript{\rm 2} CISPA Helmholtz Center for Information Security 
    \textsuperscript{\rm 3} TCS Research, Pune \\
     r.palakkadavath@deakin.edu.au,
     sarath.s@cispa.de,
\{shirish.karande,n.pedanekar\}@tcs.com

}

\begin{document}

\maketitle

\begin{abstract}
Several real-life applications require crafting concise, quantitative scoring functions (also called rating systems) from measured observations. For example, an effectiveness score needs to be created for advertising campaigns using a number of engagement metrics. Experts often need to create such scoring functions in the absence of labelled data, where  the scores need to reflect business insights and rules as understood by the domain experts. Without a way to capture these inputs systematically, this becomes a time-consuming process involving trial and error. In this paper, we introduce a label-free practical approach to learn a scoring function from multi-dimensional numerical data. The approach incorporates insights and business rules from domain experts in the form of easily observable and specifiable constraints, which are used as weak supervision by a machine learning model. We convert such constraints into loss functions that are optimized simultaneously while learning the scoring function. We examine the efficacy of the approach using a synthetic dataset as well as four real-life datasets, and also compare how it performs vis-a-vis supervised learning models.  
\end{abstract}

\section{Introduction}
Several real-life scenarios involve rating of performance on a quantitative scale using a single number. For example, employees are rated for their performance based on a number of Key Performance Indicators (KPIs) \cite{shields2015managing}. Universities are rated according to research, academic and industry performance\footnote{https://cwur.org/}. Sports teams are rated according to their competitive, branding and financial performance\footnote{https://footballdatabase.com/}. Happiness of countries is rated using factors such as GDP, life expectancy, generosity, social support, freedom and corruption\footnote{https://countryeconomy.com/demography/world-happiness-index}. Ads are rated according to click-through rate, relevance and experience as in case of Google Ad Score\footnote{https://ads.google.com/intl/en\_in/home/resources/improve-quality-score/}. Movies are rated according to their popularity, critical acclaim and financial success. Such rating systems (also called scoring functions) produce a single number to represent the performance of an observation, and serve as a concise measure easily perceived by users.

Coming up with such a scoring function involves efforts on part of those who understand the domain. Let us consider a typical scenario of rating performance of ad campaigns. Companies run ad campaigns on social media in order to increase revenue and visibility. The performance of such campaigns is measured by a number of attributes such as click rate, impressions, views, shares, follows and budget allocated for the campaign. Social media companies provide such metrics to analyze the performance of the campaign. Since there are a number of such metrics in a typical scenario (in this case, more than 50), it is difficult to present all such information to a decision maker, say the CEO of the company. In order to meaningfully represent the performance of the campaign, marketers come up with a rating on a suitable scale such that it makes quick sense to a decision maker. Such rating uses a number of performance attributes and is based on a model that intuitively makes sense to the domain experts in digital marketing. Marketers often create such a model and examine its validity by trial-and-error. They also tend to attribute quantitative importance to attributes (e.g. “Let click rate have 30\% importance, budget 20\% importance and impressions 20\% importance”). Such numbers are often ad hoc guesses, and the trial-and-error process happens to be time-consuming. Furthermore, a certain distribution of scores is needed to make sense of the score (e.g. not too many low scores, not too many high scores), which is difficult to obtain through trial-and-error. So, the outcome can often be counter-intuitive scores which experts have to live with because of the time spent in the process. 
Scenarios similar to the one outlined above exist in a variety of domains. They are often characterized by the following:

\begin{itemize}
    \item \textbf{Lack of labelled data:} Availability of labelled data in such a scenario is scarce. The scoring function often needs to rely on trial-and-error. For example, social media provide extensive data on the performance of an ad, but they rarely provide consolidated numbers. They have to be arrived at by experts.
    \item \textbf{Difficulty in labelling data:} Even if a data annotation exercise is taken up, it is difficult for domain experts to intuitively come up with a score for a given example. Such scores might differ from person to person. A pairwise comparison of examples can lead to a ranking and subsequently to a scoring scheme, but comparing n examples might require $O(n^2)$ annotations for unknown distributions \cite{ammar2011ranking}.
    \item \textbf{Individual biases:} Annotating a score may carry implicit personal biases which cannot be avoided \cite{yannakakis2015ratings, melhart2020study} .For example, rating the performance of a movie may involve the rater to be biased by genre choice. Rating the aesthetics of a photo may be affected by rater preferences of colour scheme and subjects. Such biases are implicit to the rating and are not stated explicitly in any form for them to be explained.
\end{itemize}

Given the aforementioned difficulties in crafting scoring functions, we argue that such a process can benefit from up-front incorporation of domain knowledge in an explicitly stated form, rather than hoping that it is incorporated implicitly in labelling. The process also needs to avoid unintended biases leaking implicitly and unknowingly through labelling. In our approach, domain knowledge is explicitly stated and appears as constraints on a weakly supervised model that learns a scoring function. The model incorporates these constraints as penalties or losses while learning. Experts can concur on easily observable insights (e.g. higher click-rate for an ad should lead to a higher score, or higher budget should lead to a lower score) rather than on quantitative estimates (e.g. feature weights and percentages) or more complex information (e.g. a multivariate Bayes network or a knowledge graph). So, in our approach, domain knowledge specified is neither quantitative nor complex, but is expressed as simpler functions such as monotonicity of attributes, their sensitivity, bounds on output values and the nature of output distribution. This approach is explainable as well as transparent, often a requirement for scoring. Our contribution in this paper is:  
\begin{itemize}
    \item We propose a practical label-free approach to craft a scoring model which allows incorporation of domain knowledge under a weak supervision mechanism. We convert the domain knowledge or rules into mathematical losses or penalties that can be optimized by the model. We refer to our approach as the Label-Free Weakly Supervised (LFWS) approach in this paper.
    \item We introduce two novel constraints to specify domain knowledge, viz. relative sensitivity of features and desired distribution of output scores.
    \item We demonstrate the efficacy of our LFWS approach on a synthetic dataset as well as on four real-life datasets, and compare our scoring with that given by supervised approaches.
\end{itemize}


\section{Related Work}
An explicit scoring function is often desired along with rankings in many practical scenarios. Microsoft TrueSkill \cite{10.5555/2976456.2976528}, Elo Ratings and Streetscore \cite{6910072} are examples where scores have been crafted to facilitate rankings. It is always advisable for scoring systems to provide an explanation of how the score was arrived at. Previous works like \cite{demajo2020explainable,pmlr-v84-sokolovska18a} incorporate explainability into scoring systems by combining multiple AI techniques to cater differently for each of the features, learning binning strategies on the features, and perturbing the features so as to compare the variation in the score. 

Additionally, there are prior works~\cite{Gupta_Marla_Sun_Shukla_Kolbeinsson_2021, NEURIPS2018_caa20203} that provide a framework to model the shape relationship between the input features and the output as mathematical constraints. These constraints are added as regularization terms along with the original regression or classification objective. However, the aforementioned work requires label information to perform modelling. 

When it comes to unsupervised or weakly supervised paradigms for ranking, prior works like  \cite{dehghani2017neural,xu2019passage,10.1007/978-3-642-24955-6_72} usually aggregate opinions from multiple weak supervision signals of ranks to perform ranking. \cite{alattas2019unsupervised} is an unsupervised scoring technique that calculates the weights of each feature based on the similarity in their correlation coefficients (feature clustering). Another recently proposed unsupervised scoring technique \cite{ichikawa2020unsupervised} is applicable only for binary features.

In our LFWS approach, we allow incorporation of domain expertise into a learning model as a weak supervision signal. We approach the problem as optimizing a neural network with constraints \cite{madry2019deep,DBLP:journals/corr/abs-2002-10742}. Unlike existing methods, we do not have any label information with us other than the domain constraints. One way to inject domain level constraints into a machine learning model is to include it as a loss or penalty to the model \cite{10.5555/3298483.3298610,chen2018neural,raissi2017physics,fischer2019dl2,DILIGENTI2017143,Gupta_Marla_Sun_Shukla_Kolbeinsson_2021}. But these are usually physical constraints in equations or logical constraints expressed in first order logic or shape constraints between the input and the output. Our LFWS approach includes business rules in the model by converting them to specific objective functions that the model can optimize. 

\section{Crafting Scoring Functions}

Consider assigning scores to a set of ad campaigns based on the performance metrics provided by the social media company, such as clicks, impressions, total amount spent and leads generated. In our approach, instead of labels, domain experts specify constraints such as:
\begin{itemize}
    \item Feature $leads$ $generated$ should contribute the most to the score
    \item Score should be bounded between $a$ and $b$ 
    \item $x\%$ of ads should be given a score lower than $y$
\end{itemize}

The aforementioned domain constraints can be translated to identifying an abstract function $f$ with the help of its certain mathematical properties, such as relative values of its gradients (sensitivity), bounds on the derivative (smoothness) and constraints on its range set (bounds or output constraints on score). We model the problem as regularized constrained optimization over all functions $f:\mathbb{R}^n$ $\xrightarrow{}$ $\mathbb{R}$ parametrized by $\theta$, the weights and biases of a specified neural network, such that $f(x)$ represents the score for a set of observed features $x$. The given constraints are then converted to losses that $f$ can optimize. In particular, we have formulated the following losses to allow incorporation of domain expertise: 

\subsection{Monotonicity Constraint}
The fundamental constraint to our formulation is the monotonicity constraint. In some cases, experts may not be able to completely specify the other constraints for the model to reach a single solution, and without a supervision loss, multiple solutions may arise for the same specification of constraints. We address this using the monotonicity constraint, as it makes up for the absence of supervision loss and provides a good amount of regularization in the neural network \cite{eberle2021monotonicitybased}.

In practice, data features contribute to a score in a positive or a negative way. In the case of rating movies, features such as gross income, meta score and votes intuitively affect the score positively. However, the budget might have a negative effect on the score. In some cases, a feature such as movie length can contribute negatively to the score when its values are at extremes and positively otherwise. Such knowledge of trends is pivotal to craft the scoring function without labels, but can be easily obtained from domain experts.

In order to incorporate this knowledge, we propose a monotonic neural network that takes in the input and predicts a score proportional to the feature values. The input is normalized to lie in [0,1]. We use a feature engineering step that transforms negatively affecting input features so as to make them invert their initial proportions to the scores. For example, if the feature is originally $x$, and is supposed to have a negative effect to the score, we transform it to $1-x$. If $x$ is supposed to affect the score positively at extremes and negatively in between (convex shape), it may be transformed piece-wisely to $x^{2}$ or $x$ in the first half and $(1-x)^2$ or $1-x$ in the second half. These transformations need not exactly represent the relation in the data, and can be informed guesses that the model use as informative prior.

To create a monotonic network, 
following \cite{10.5555/302528.302767}, we first let the network learn weights in the log domain during optimization. At the time of forward pass, we encompass the exponential function over the weights reversing the log-weights. This ensures that the weights are positive, ensuring that the network is monotonic. 

\subsection{Boundedness Constraint}
One of the common constraints is to bound the score between a given range, [a,b]. The most obvious choice for this task seems to be an appropriate activation function at the output layer of the network. However, we observed it to be a restrictive choice, since the scores did not spread much within the bounded region and resulted in isolated modes. Moreover, we also found it difficult to optimize the other constraints simultaneously. Hence, we use a modified version of the hinge loss function to bound the scores in [a,b]. 
\begin{align}
\textit{BL} = \max(0,a - f(x)) + \max(0, f(x) - b)
\end{align}
In some cases, the output may not be completely distributed over the range [a,b]. Without loss of information from other losses, we scale up the scores to completely occupy [a,b] in that case. If additional information regarding the modes of the scores is available, a mode loss $ML$ can be used to increase the density of the scores at a particular score, $m$.
\begin{align}
\textit{ML} = \max(0, \mid m - f(x) \mid) 
\end{align}

\subsection{Sensitivity Constraint}
In practice, a domain expert should be able to provide the relative importance of features deemed important to the score. For example, for an ad effectiveness score targeted at generating leads, the number of leads generated should contribute to the score more than other features deemed important such as click-through rate, impressions and cost. Furthermore, one can specify the click-through rate to be more important that the remaining important features. This gives flexibility for the user to stress on a specific set of input features in making a larger impact on the prediction, without providing explicit weights for their importance. 

While works such as \cite{cheng2016wide,al2011empirical} solve this in a supervised setup, we implement it using a novel sensitivity loss. We utilize the model's susceptibility against each of its inputs by leveraging its sensitivity. Given $f$ and input $x$, the sensitivity of the model to an input feature $i$ is the rate of change of the output $f(x)$ with respect to change in $x_{i}$, $\frac{\partial f(x)}{\partial x_{i}}$. In order to emphasize $i$ in $f(x)$, we formulate a loss $SL$ that favours the sensitivity of $i$ over any other feature, say $j$ where $i \neq j$.
\begin{align}\label{eq:3}
\textit{SL} =  \frac{\sum_{j}{\frac{\partial f(x)}{{\partial x_{j}}}}}{\frac{\partial f(x)}{\partial x_{i}}}    
\end{align} 
We extend the above to include the importance of multiple features according to their priorities. Let us consider that feature $i$ is the most important feature, then our sensitivity loss will be as displayed in Equation \ref{eq:3}, where $j$ refers to every other feature except $i$. In case there were other important features, say $k$ and $m$ with equal importance, but less important than $i$, then an extra term is added to the sensitivity loss with with the gradients of $f$ with respect to $j \neq i,k,m$ in numerator and $k,m$ in denominator. We use the automatic differentiation package in PyTorch \cite{paszke2017automatic} to implement this loss.

\subsection{Output Constraint}
The distribution of the output score is an important consideration in its interpretation. An output constraint can be used to enforce scores, distributed randomly at first, to follow a target probability distribution. For example, a Gaussian distribution is an apt choice to rate employee performance as often referred to as the Bell Curve \cite{stewart2010forced}. There are scoring schemes such as sports player rankings, competitive exam scores and interview scores that favour a right skewed distributions such as an exponential distribution. Also, rather than providing an entire distribution, one can impose a constraint on the quantiles of the score distribution.

The scores learned by our model initially follow a random distribution ($q$). Instead, we require it to conform to a target distribution ($p$). So, we add a loss that computes the similarity (K-L or J-S divergence) between distributions $p$ and $q$ as the output constraint. 
Before calculating the loss, we need to approximate the initial density function of $q$. 
We chose to approximate this $q$ in a parametric form, whose parameters are the functions of the output of the network taken while training: mean $\mu_{1} = mean(f(x))$ and standard deviation $\sigma_{1} = std(f(x))$ where $x$ is the batch input to the network and $\mu$ and $\sigma$ are calculated over mini-batches of size 64. It is possible to end up with a better approximation by considering higher-order moments of the output distribution. However, this representation is sufficient for our current requirement as the information is provided by a domain expert who is not likely to be aware of the higher-order moments. But in the future, more generic approaches can be explored to approximate the score distribution. The target distributions $(p)$ that the scores need for follow are listed below with their respective losses.

\subsubsection{Gaussian Distribution}
We enforce the scores to follow a normal distribution ($p$) with desired parameters (specified by a domain expert): mean $\mu_{2}$ and variance $\sigma_{2}$: $p \sim \mathcal N(\mu_{2}, \sigma_{2}^2)$. We minimize KL Divergence between the given distribution $(p)$ and the current distribution of the score $(q)$ by modifying $q$'s parameters, $\mu_{1}$ and $\sigma_{1}$. We chose the zero-forcing reverse KL loss as $q$ is a Gaussian and and it is easier to evaluate the loss in closed form with most of the target distributions. It also ensures that the density of the final scores, $q$ where the target $p=0$ to be zero as well. 
\begin{equation}
\begin{aligned}
 \textit{KL(q, p)} &= \EX_{q(x)}\log q(x) -\EX_{q(x)} \log p(x)\\
&= \log \frac{\sigma_2}{\sigma_1} + \frac{\sigma_1^2 + (\mu_1 - \mu_2)^2}{2 \sigma_2^2} - \frac{1}{2}    
\end{aligned}
\end{equation}
\subsubsection{Exponential Distribution}
If most of the scores lie in the low ranges, we choose $p$ to be exponential with a distribution parameter $\lambda$: $p \sim \text{Exp}(\lambda)$. The KL-divergence for the same is below.
\begin{equation}
\begin{aligned}\label{eq:5}
 \textit{KL(q, p)} 
&= -\frac{1}{2} - \frac{1}{2}\log (2\pi\sigma_{1}^2) -\log \lambda +\mu_{1}\lambda   
\end{aligned}
\end{equation}

\begin{table*}
  \centering
  \begin{adjustbox}{width=1\textwidth}
    \begin{tabular}{ccccccccccccc}
    \toprule
    \textbf{Monotonicity?} & \textbf{Case} & \textbf{Constraint} & \textbf{Rank Correlation} & \textbf{RMSE} & \textbf{KL divergence} & \textbf{Min} & \textbf{Max} & \textbf{\% in bound} & \multicolumn{4}{c}{\textbf{Feature Correlation (x1,x2,x3,x4)}} \\
    \midrule
    \multirow{2}{*}{} & 1     & Ground Truth & 1     & 0     & 0     & 19.62 & 654.45 & 100   & 0.02  & 0.18  & 0.58  & 0.76 \\
          & 2     & XGB - Supervised & 0.99  & 12.88 & 0.013 & 71.48 & 609.79 & 100   & 0.02  & 0.15  & 0.58  & 0.76 \\
    \midrule
    \multirow{7}{*}{No} & 3     & Distribution (Gauss KL) & 0.67  & 63.41 & 0.0116 & 127.9 & 516.93 & 100   & 0.42  & -0.28 & 0.64  & 0.42 \\
          & 4     & Bound & 0.75  & 276.32 & 682.6 & 27.69 & 77.02 & 100   & 0.51  & 0.48  & 0.46  & 0.45 \\
          & 5     & Sensitivity & -0.41 & 373.54 & 1073.56 & -71.4 & -19.31 & 0     & -0.69 & -0.5  & -0.45 & -0.03 \\
          & 6     & Bound + Distribution & 0.73  & 57.49 & 0.00011 & 126.6 & 540.09 & 100   & -0.1  & 0.23  & 0.84  & 0.25 \\
          & 7     & Distribution + Sensitivity & -0.07 & 8077.2 & 4.143 & -22169 & 45896 & 0.028 & 0.35  & -0.76 & -0.07 & 0.14 \\
          & 8     & Bound + Sensitivity & 0.29  & 106.61 & 0.0606 & -66.33 & 695.57 & 99.9  & 0.57  & 0.08  & -0.3  & 0.56 \\
          & 9     & All losses together & 0.36  & 129.94 & 0.244 & 37.76 & 2332.515 & 99.4  & 0.4   & -0.06 & -0.09 & 0.51 \\
    \midrule
    \multirow{7}{*}{Yes} & 10    & Distribution (Gauss KL) & 0.2   & 99.58 & 0.000978 & 56.88 & 601.04 & 100   & 0.2   & 0.97  & 0.01  & -0.01 \\
          & 11    & Bound & 0.79  & 144.5 & 13.78 & 76.85 & 276.18 & 100   & 0.45  & 0.49  & 0.49  & 0.48 \\
          & 12    & Sensitivity & 0.77  & 12422698 & 16.03 & 0     & 25355356 & 0     & 0.01  & 0.02  & 0.03  & 1 \\
          & 13    & Bound + Distribution & 0.69  & 60.97 & 0.0002 & 0.49  & 556.68 & 99.9  & 0.01  & 0.12  & 0.99  & 0.11 \\
          & 14    & Distribution + Sensitivity & 0.8   & 50.63 & 0.0135 & 4.73  & 646.92 & 99.9  & 0.01  & 0.02  & 0.08  & 1 \\
          & 15    & Bound + Sensitivity & 0.79  & 69.22 & 0.113 & 0.06  & 738.09 & 99.9  & 0.01  & 0.02  & 0.07  & 1 \\
          & 16    & All losses together & 0.9   & 33.37 & 0.0035 & 61.16 & 602.18 & 100   & 0.01  & 0.01  & 0.29  & 0.95 \\
    \bottomrule
    \end{tabular}%
    \end{adjustbox}
    \caption{Comparison of results with scores generated with and without monotonicity constraints and different losses to ground truth and supervised learning}
  \label{tab:addlabel}%
\end{table*}%


\section{Experimental Details}
\subsection{Datasets}

We use one synthetic dataset to demonstrate how the proposed losses in our model affect the output scores. In addition, we use three real-life datasets to compare our results with those of a supervised approach. We use an additional dataset that does not have a scoring ground truth and we use it to simulate the process to be followed in practice.  The datasets are as follows:

\subsection{Synthetic Dataset}
We create a synthetic dataset using four features sampled independently from a Gaussian distribution (mean = 10, variance = 3). There are a total of 10,000 data points, split to test and train with 70\% and 30\% weightage respectively. The score $y$ is a combination of the 4 features given by:
$y = x_{1} + a_{1}*x_{2} + a_{2}*x_{3} + x_{4}^{2}$
where $x_{i}$s are the input features and $a_{1} = 5$ and $a_{2} = 15$.

\subsubsection{IMDB Dataset}
This dataset is a collection of roughly 6000 movies (US releases) from 1930 - 2019\footnote{https://www.kaggle.com/stefanoleone992/imdb-extensive-dataset}. These movies are described by number of votes cast, average votes, budget, gross income, meta-score, and number of user / critic reviews. We use the CPI library\footnote{https://github.com/palewire/cpi} to account for the inflation rate for gross income and budget. Budget is the only feature that is deemed negatively correlated with the score. The score is bounded to [0,10].

\subsubsection{CWUR Dataset}
This dataset comes from the Centre for World University Rankings\footnote{https://cwur.org/}. It consists of the following features: national rank, quality of education, alumni employment, quality of faculty, publications, influence, citations, broad impact and patents. Since the true college ratings are available for this dataset, we use it for comparison with supervised models in section \ref{sec5}. The scores lie between: $[40,100]$. CWUR website mentioned that 40\% of the weight-age was given to research performance: a combination of patents, publication, influence and citation. So we consider these as the important features for this dataset.

\subsubsection{Journal Rank Dataset}
This dataset contains the impact factor ratings for the top 1000 journals in 2019\footnote{https://www.kaggle.com/umairnasir14/impact-factor-of-top-1000-journals}. The scores are based on: \% cited compared to the previous year, source normalized impact per paper (SNIP), SCImago journal rank (SJR: measure of the scientific influence of scholarly journals). The impact rating is between [8,150]. All the features contribute positively to the score with the feature SJR having the most impact on the score.

\subsubsection{Ad Campaign Dataset}
This proprietary dataset contains information about 450 ads that were targeted at a lead generation campaign. The data was collected from LinkedIn ad campaigns of a multi-national company. The attributes are: amount spent, clicks, click through rate, cost per click, cost per lead, impressions, leads generated and lead generation rate etc. According to domain experts, while amount spent, cost per click, cost per lead affect the score in a negative way, the rest of the attributes contribute positively. The scores are bounded between [0,10]. Form fill rate, cost per lead and the combination of impressions, no of leads and clicks are considered as the set of important features in that order. This dataset does not have ground truth scores and we use it to simulate the process followed while using our approach.

\begin{table*}
\centering
\begin{adjustbox}{width=1\textwidth}
\begin{tabular}{ccccccccc}
\toprule
     \multicolumn{3}{c}{IMDB}  &  \multicolumn{3}{c}{CWUR} & 
     \multicolumn{3}{c}{Ad}\\
    \midrule
    Features  & \multicolumn{2}{c}{$\rho$} & 
    Features  &  \multicolumn{2}{c}{$\rho$} & 
     Features  &  \multicolumn{2}{c}{$\rho$} \\
 & Without Sen & With Sen &  & Without Sen & With Sen & & Without Sen & With Sen\\
 \midrule
\cellcolor[HTML]{b8b8b8} Gross Income & \cellcolor[HTML]{b8b8b8}-0.154 & \cellcolor[HTML]{b8b8b8}0.480 & National Rank & 0.603 & 0.590 & \cellcolor[HTML]{b8b8b8} Form Fill Rate & \cellcolor[HTML]{b8b8b8}0.480 & \cellcolor[HTML]{b8b8b8}0.887 \\
1-Budget & 0.503 & 0.202 & Quality of Education & 0.515 & 0.522 & \cellcolor[HTML]{b8b8b8} Combined & \cellcolor[HTML]{b8b8b8} 0.582 & \cellcolor[HTML]{b8b8b8} 0.855 \\
Meta Score & 0.745 & 0.249 & Alumni Employment & 0.352 & 0.367 & \cellcolor[HTML]{b8b8b8} 1-Cost Per Lead & \cellcolor[HTML]{b8b8b8}0.511 & \cellcolor[HTML]{b8b8b8}0.880 \\
User Review & 0.101 & 0.174 & Quality of Faculty & 0.564 & 0.570 & Leads & 0.435 & 0.854 \\
Critics Review & 0.211 & 0.059 & \cellcolor[HTML]{b8b8b8} Publication, Influence, Citation 
 & \cellcolor[HTML]{b8b8b8}0.807 & \cellcolor[HTML]{b8b8b8}0.833 & Click Rate & 0.402 & 0.190 \\
Votes & 0.721 & 0.248 & Patents & 0.815 & 0.834 & Clicks & 0.493 & 0.630 \\
Average Votes & 0.145 & 0.239 & Broad Impact & 0.457 & 0.472 & Impressions & 0.266 & 0.679 \\
 \hline
\end{tabular}
\end{adjustbox}
\caption{Spearman's rank correlation values between input features and generated scores}
\label{table:corrtable}
\end{table*}

\begin{figure*}[ht]
  \begin{subfigure}{0.23\textwidth}
  \includegraphics[width=\linewidth]{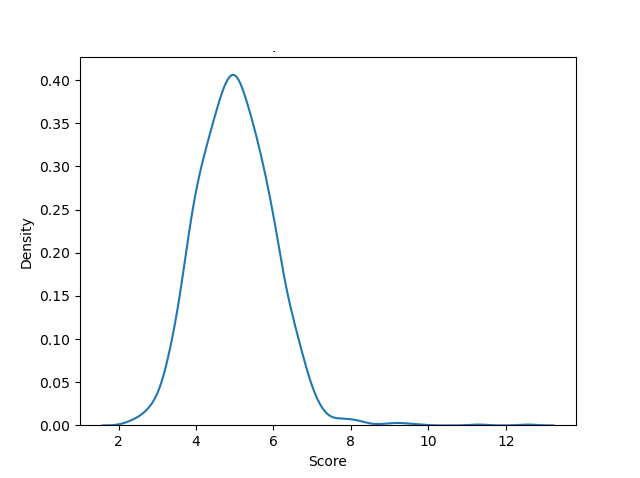}
  \caption{IMDB - Gaussian}
  \label{fig:gs_imdb}
  \end{subfigure}
    \begin{subfigure}{0.23\textwidth}
  \includegraphics[width=\linewidth]{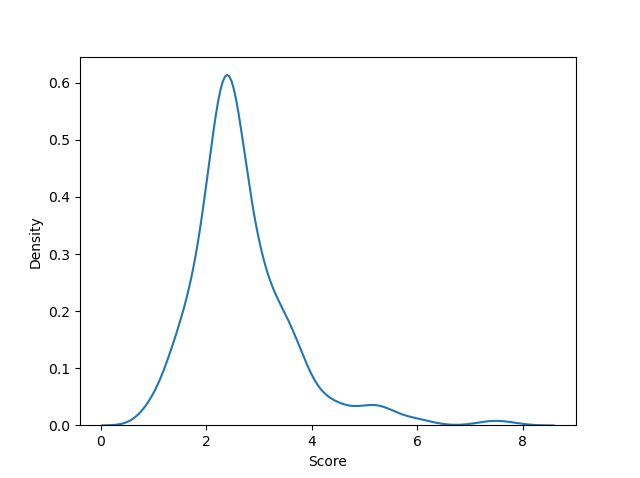}
  \caption{Ad - Gaussian}
  \label{fig:gs_ad}
  \end{subfigure}
  \begin{subfigure}{0.23\textwidth}
  \includegraphics[width=\linewidth]{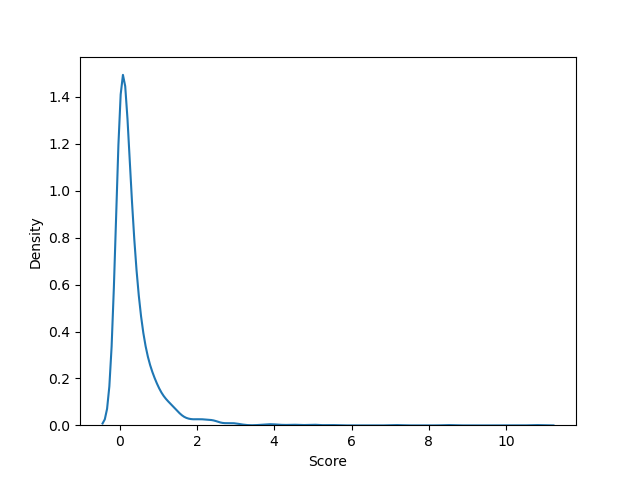}
  \caption{IMDB - Exponential}
  \label{fig:es_imdb}
  \end{subfigure}
    \begin{subfigure}{0.23\textwidth}
  \includegraphics[width=\linewidth]{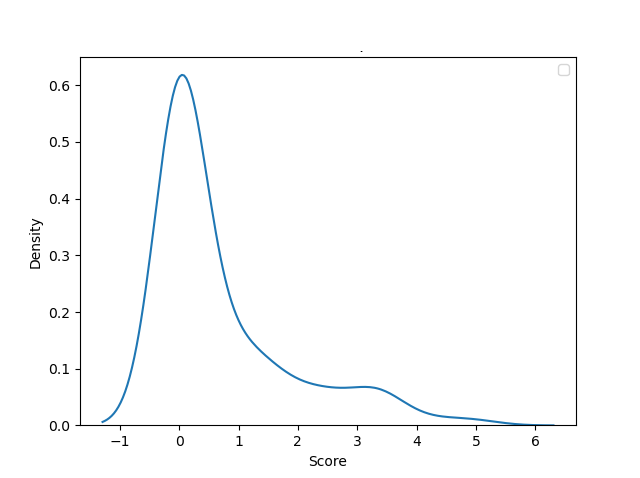}
  \caption{Ad - Exponential}
  \label{fig:es_ad}
  \end{subfigure}
    \caption{Kernel density estimates of generated scores}
    \label{fig:exp}
\end{figure*}

\subsection{Experimental Set Up}
For our experiments, we specified $f$ as a 3-layer dense neural network with ELU nonlinearity for the intermediate layers. We used the dataset attributes as input to the network. The final layer activation was linear and provided the score as an output. The objective of the model was to minimize a weighted sum of the constraint losses described above. The weights [$\alpha, \beta, \gamma$] are hyper-parameters whose values are decided based on the priority of the constraints.
\begin{align}\label{finalloss}
\textit{$L_{f}(x)$} = \min_{f}(\alpha * \textit{BL} + \beta * \textit{SL} + \gamma * \textit{KL})
\end{align}
In general, we used a value of 1 for $\alpha$, $\beta$ and $\gamma$. It is possible to give higher values (say, 10) to $\alpha$ as bound losses are typically higher, and lower values (say, 0.1) to $\beta$ as sensitivity losses are typically lower. For each experiment, we randomly split each dataset into two parts. 70\% of the data was used to train the LFWS model (without labels and with weak supervision of constraints). We used the remaining 30\% of the data to calculate scores using the trained model. We examined how well the generated scores adhered to the constraints and analyzed how the proposed losses affected the crafted scores. The results of this analysis are presented in the following section. 

\begin{table*}
\centering
\begin{adjustbox}{width=1\textwidth}
\begin{tabular}{ccccccccccccccccccc}
\toprule
\multicolumn{1}{c}{}  &  
\multicolumn{11}{c}{CWUR Data Set} &
\multicolumn{7}{c}{Journal Data Set}\\
\midrule
\multicolumn{1}{c}{Model}  &  \multicolumn{7}{c}{$\rho$(score,feature)} &
     \multicolumn{1}{c}{Rank}&
     \multicolumn{1}{c}{\multirow{2}{*}{RMSE}}&
     \multicolumn{1}{c}{\% in} &
     \multicolumn{1}{c}{(min, max)} &
     \multicolumn{3}{c}{$\rho$(score,feature)} &
     \multicolumn{1}{c}{Rank}&
     \multicolumn{1}{c}{\multirow{2}{*}{RMSE}}&
     \multicolumn{1}{c}{\% in }&
     \multicolumn{1}{c}{(min, max)}\\
\cmidrule(lr){1-1}     
\cmidrule(lr){2-8}
\cmidrule(lr){13-15}
  & $x_{1}$ & $x_{2}$ & $x_{3}$ &
$x_{4}$ & $x_{5}$ & $x_{6}$ & $x_{7}$ & corr & & bound & bound & $x_{1}$ & $x_{2}$ & $x_{3}$ & corr & & bound & bound\\
\cmidrule(lr){2-8}
\cmidrule(lr){9-9}
\cmidrule(lr){10-10}
\cmidrule(lr){11-12}
\cmidrule(lr){13-15}
\cmidrule(lr){16-16}
\cmidrule(lr){17-17}
\cmidrule(lr){18-19}
  Ground Truth  & 0.26 & 0.62 & 0.91 & 0.88 & 0.84 &
 0.94 & 0.61 & 1.0 & 0 & 100 & (44.2, 92.8) & 0.35& 0.64& 0.78& 1.0 & 0 & 100 & (8.8, 123.7)\\
 XGBoost  & 0.27 & 0.61 & 0.92 & 0.87 & 0.84 &
 0.94 & 0.57 & 0.98 & 1.65 & 100 & (43.3, 85.4) & 0.42& 0.75 & 0.81& 0.77 & 8.85 & 100 & (9.2, 59.8)\\
 Supervised NN & 0.25 & 0.01 & 0.10 & 0.20 & 0.16 &
 0.13 & 0.04 & 0.12 & 7.50 & 100 & (31.3, 87.1) & 0.94& 0.30 & 0.23& 0.42 & 10.52 & 100 & (17.08, 17.71) \\

  LFWS & 0.48 & 0.61 & 0.94 & 0.86 & 0.84 &
 0.93 & 0.53 & 0.90 & 17.6 & 78 & (11.6, 85.9) & 0.63& 0.59& 0.79& 0.78 & 12.92 & 99 & (2.5, 60.24) \\ 
\bottomrule
\end{tabular}
\end{adjustbox}
\caption{Comparison of LFWS scores with those learnt using supervised models}
\label{table:supertable}
\end{table*}

\begin{figure}[ht]
    \centering
  \begin{subfigure}{0.23\textwidth}
  \includegraphics[width=\linewidth]{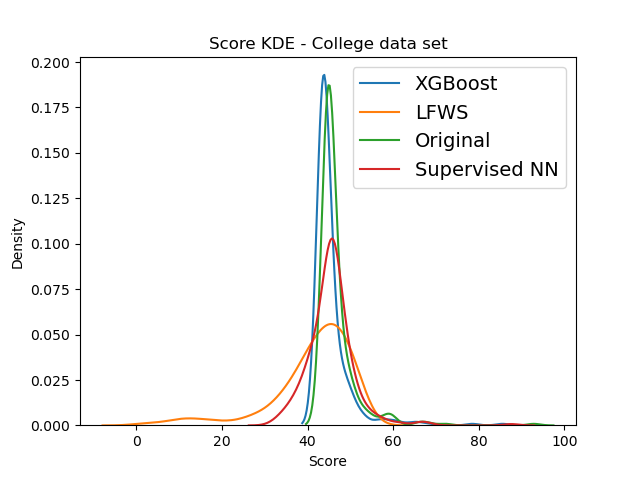}
  \caption{CWUR Data Set}
  \label{fig:ds_j_og}
  \end{subfigure}
  \begin{subfigure}{0.23\textwidth}
  \includegraphics[width=\linewidth]{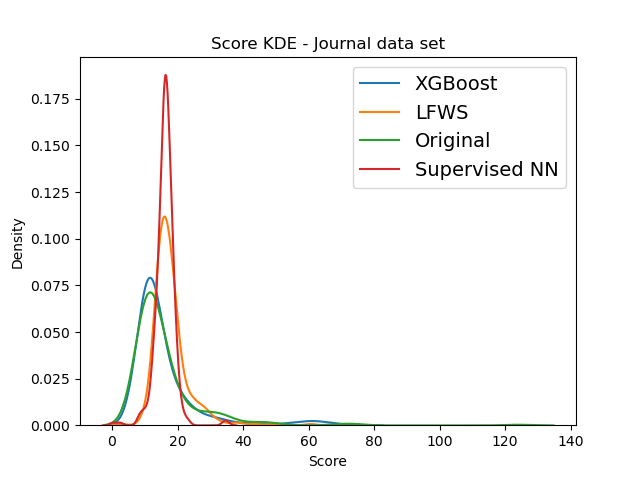}
 \caption{Journal Rank Data Set}
  \label{fig:ds_j_xb}
  \end{subfigure}
    \caption{Kernel density estimates for generated scores}
    \label{fig:jr}
\end{figure}

\section{Results and Discussion}
\subsection{Synthetic Dataset}
The results on the experiments with the synthetic dataset are summarized in Table \ref{tab:addlabel}. The columns indicate performance metrics, viz. Spearman's rank correlation that evaluates a monotonic relationship between two variables, root mean squared error, KL divergence with ground truth distribution, minimum and maximum values of score, percentage of values within bounds and correlation of the score with each of the four features. The top row (case 1) indicates the ground truth of the scoring function, while the next row (case 2) shows results using an XGBoost supervised model with ground truth as labels. The next two sets of values indicate performance metrics for our model without (cases 3-9) and with (cases 10-16) monotonicity constraints, respectively. Each set also shows how different combinations of losses affect the performance metrics. We observed that the supervised model performed well as expected on the synthetic dataset and was able to learn the scoring function. The set with monotonicity constraints in general outperformed the other set in all metrics. Even with all the losses taken together (case 9), the set without montonicity constraints could not learn the scoring function well. We made the following observations with respect to the set with monotonicity constraints:
\begin{itemize}
    \item Having only the distribution constraint generated (case 10) a score population reasonably within bounds and with low KL divergence, but the overall and feature correlations could not be learnt. 
    \item Only the bound loss (case 11) imposed bounds within the given range, but could not cover the whole range, as the loss over the learnt range was still zero. While doing so, it seemed to create a score that was equally sensitive to all features.
    \item Considering only the sensitivity loss (case 12) was not able to control the bounds and the population, thus not allowing learning to happen even on feature sensitivity.
    \item When only the sensitivity loss was omitted (case 13), the model generated a reasonable population but could not learn feature sensitivity. 
    \item When only the bound loss was omitted (case 14), it still generated a population within bounds, perhaps owing to the distribution constraint, but could not learn feature sensitivity. A similar observation was made when only the distribution loss omitted (case 15). 
    \item When all losses were considered together (case 16), we observed reasonable values for overall correlation, range and feature correlations. 
\end{itemize}
Though the learning of feature correlations was not as good as that by the supervised model, the relative sensitivity rank was maintained. We found that we could get improved feature correlations over case 16 by adjusting the weights on sensitivity losses for individual feature by one order of magnitude (from 1 to 10). But we noted that such tweaking of weights is not appropriate in the absence of ground truth, and can in future be performed automatically based on expert feedback on simply whether the model has learnt enough sensitivity to a given feature. We believe that the experiment on synthetic dataset shows the capability of the model to learn without labels when all proposed losses are considered.  

\subsection{Real-life Datasets}
\subsubsection{Sensitivity Loss}
We tested the performance of sensitivity loss on the IMDB and Ad datasets. We measured the efficacy of the sensitivity loss with the help of Spearman's rank correlation coefficient $\rho$. Positive values of $\rho$ indicate a monotonically increasing relationship, while negative values of $\rho$ indicate a monotonically decreasing relationship between the inputs and the score. Table \ref{table:corrtable} displays the correlation of the input features with the output score by performing ablation on the sensitivity loss, keeping all other losses as they are. We observed that for every important feature in each of the datasets (shaded in grey), $\rho$ increased significantly with the addition of the sensitivity loss. Also, important features had the highest values for $\rho$ after the addition of the sensitivity loss. This shows that the sensitivity loss is successful in making the important features influence the score. 

\subsubsection{Output Loss}
We used kernel density estimation plots on the scores given by the model to show how output losses aid in crafting an output score distribution. Figure \ref{fig:exp} shows the predicted scores of IMDB (a, c) and Ad (b, d) data sets being forced to fit a Gaussian curve and exponential curve respectively. The KL Divergence was reduced between the original score distribution $(q)$ and a target distribution (Gaussian $(p)$ with mean $\mu_{1}$=(5,5) and standard deviation $\sigma_{1}=(1,1))$ in the case of figures (a) and (b) in \ref{fig:exp}. In case where the target distribution was exponential, $\lambda$ value of 1 was chosen for both (c) and (d). Figure \ref{fig:exp} displays the kernel density plot of the scores obtained after training the data using \ref{eq:5}. We observed that the scores followed the target distributions. Some of the scores exceeded the bounds, probably due to multiple divergent constraints. However, density of such points was low. This can be corrected by changing the importance of the distribution loss in the model, if needed.

\subsubsection{Comparison with supervised models}\label{sec5}
We compared the scores generated with the LFWS approach to those with supervised regression models, viz. XGBoost and a 3-layer supervised neural network with the same network architecture as that of the LFWS model. We used the ground truth information from the CWUR dataset and Journal Rank dataset as labels for the supervised case. For the LFWS loss construction, the following constraints were used using domain expertise already available at the websites hosting the data:\\
\\
\textit{CWUR data set}
\begin{itemize}
    \item Importance of features: patents(x6) = publications (x3) $>$ influence (x4) = citation (x5) $>$ quality of faculty (x2) = broad impact (x7) $>$ national rank (x1)
    \item Bound on scores: - [40, 100] with a mode around 45
\end{itemize}
\textit{Journal rank data set}
\begin{itemize}
    \item Importance of features: SJR (x3) $>$ SNIP (x2) $>$ \% cited (x1)
    \item Bound on scores: - [5, 150] with a mode around 13
\end{itemize}
We added the following losses to the LFWS model for either dataset:
\begin{itemize}
    \item Sensitivity loss was added for all the features above in the order of their importance.
    \item Bound loss was added for both lower and upper bound scores. It was slightly modified by adding a square function (by maintaining the original sign) over the difference of $f(x)$ and the upper/lower bounds. This gave a larger penalty to the scores that didn't satisfy the bound loss. 
    \item A mode loss was added to focus most of the scores towards the mode. 
\end{itemize}
Table \ref{table:supertable} gives the comparison of results of our LFWS model with those of XGBoost, a 3-layer supervised neural network model along with the ground truth values. It shows the Spearman's rank correlation coefficient, $\rho$(original score, features), between the original score (ground truth labels) and each of the features ($x1-x7$, $x1-x3$) in the first row. The subsequent rows show the same between the calculated scores by various methods and the features, $\rho$(predicted scores, features). We also calculated the rank correlation in a similar way which compares the ranks of scores calculated for examples with those for the ground truth. We compared the range of the calculated scores to that of the ground truth, and also found the percentage of scores generated within the stipulated bounds. We made the following observations about this experiment:
\begin{itemize}
    \item Supervised XGBoost came closest to the ground truth for both the datasets as expected. But the LFWS model could capture the correlation information between the score and the features quite well for these datasets without any label information.
    \item XGBoost also showed the highest rank correlation to the true scores with 0.98 while the LFWS scoring model was close behind with a rank correlation of 0.90 for CWUR dataset. A similar trend followed for the Journal rank dataset with a rank correlation of 0.81 for XGBoost model and 0.76 for the LFWS scoring model. 
    \item XGBoost seemed to perform well on the CWUR dataset, but not so well on the Journal rank dataset. Our models faltered with respect to the lower bound, but in every other case, they did almost as well as XGBoost. Even with the lower bound exceeded, almost 80\% and 99\% of the scores lay within the defined bound for the CWUR and Journal rank datasets respectively.
    \item Figures \ref{fig:ds_j_og} and \ref{fig:ds_j_xb} show KDE plots for the scores of CWUR and Journal rank datasets respectively. For both the datasets, we observed that the modes specified by the constraints (45,13) were followed in the LFWS scoring model.
\end{itemize}

\section{Conclusion and Future Work}
In this paper, we presented a novel label-free approach to crafting scores from multivariate data with the help of weak supervision information in form of constraints. We did so by introducing two novel losses in the learning model to allow specification of relative sensitivity of features to scores and distribution of output as constraints. We also incorporated monotonocity and bound constraints in the learning model. We explored the efficacy of the model in incorporating domain expertise as constraints and in generating desired scores using a variety of datasets. We also showed that the LFWS score rankings were close to supervised regression models such as XGBoost, even if there was no label information. 

The comparison with supervised models is presented only to show the efficacy of our approach. In practice, we assume that no such labels are available and the validation of the scores has to be carried out by domain experts. Though not reported explicitly in this paper, we found that we could improve the results of our model by changing hyperparameters such as loss weights for individual losses and sensitivity losses. But such tweaking leads to a trial-and-error approach similar to that described in the introduction. In future, we would like to incorporate such tweaking automatically based on expert feedback based on simple observations about the behaviour of the output. We would also like to develop low-effort validation methods for experts such as pairwise comparisons and visualization-driven tweaking of scoring functions. In fact, our immediate goal is to test the real-life performance of the LFWS approach in the hands of domain experts to craft an ad effectiveness score using the Ad dataset mentioned in the paper. 

We believe that the proposed approach is a practical tool for domain experts to specify what they know in a simple form and generate desired scoring functions. They do not have to label the data explicitly and do not have to come up with empirical models or ad hoc weightages. Since the constraints are stated explicitly, the domain knowledge is explainable and the risk of implicit biases is avoided. 



\bibliography{aaai23.bib}
\end{document}